\documentclass[11pt,draftcls,onecolumn]{IEEEtran}

\usepackage{amsmath,graphicx}
\usepackage{epsfig}
\usepackage[center]{caption}
\usepackage{epstopdf}
\usepackage{array}
\usepackage{subfig}
\usepackage{multicol}
\usepackage{turnstile}
\usepackage{threeparttable}
\usepackage{longtable}
\usepackage{fancyvrb}
\usepackage{changepage}
\usepackage[stable]{footmisc}
\usepackage[T1]{fontenc}
\usepackage{color}
\input xy
\xyoption{all}
{\begin{list}{}%
         {\setlength{\leftmargin}{#1}}%
         \item[]%
}
{\end{list}}

\title{A Hand-Held Multimedia Translation and Interpretation System with Application \\to Diet Management}
\author{Albert Parra*~\IEEEmembership{Student Member,~IEEE}, Andrew W. Haddad~\IEEEmembership{Student Member,~IEEE}, \\Mireille Boutin~\IEEEmembership{Member,~IEEE} and Edward J. Delp~\IEEEmembership{Fellow,~IEEE}

\thanks{This work was partially supported by the U.S. Department of Homeland Security's VACCINE Center under Award Number 2009-ST-061-CI0001. Address all correspondence to M. Boutin (mboutin@ecn.purdue.edu).}

\thanks{The authors are with the Department of Electrical and Computer Engineering, Purdue University, West Lafayette, IN 47907 (e-mail: aparrapo@ecn.purdue.edu; haddada@purdue.edu; mboutin@purdue.edu; ace@ecn.purdue.edu).}

\thanks{EDICS categories: 1-COMP, 6-MEDS, 8-WMMM, 9-MASI}
}

\begin{document}
\maketitle


\begin{abstract}
\label{abstract}
We propose a network independent, hand-held system to translate and disambiguate foreign restaurant menu items in real-time. The system is based on the use of a portable multimedia device, such as a smartphones or a PDA. An accurate and fast translation is obtained using a Machine Translation engine and a context-specific corpora to which we apply two pre-processing steps, called translation standardization and \textit{$n$-gram consolidation}.
The phrase-table generated is orders of magnitude lighter than the ones commonly used in market applications, thus making translations computationally less expensive, and decreasing the battery usage.
Translation ambiguities are mitigated using multimedia information including images of dishes and ingredients, along with ingredient lists. We implemented a prototype of our system on an iPod Touch Second Generation for English speakers traveling in Spain. Our tests indicate that our translation method yields higher accuracy than translation engines such as Google Translate, and does so almost instantaneously.
The memory requirements of the application, including the database of images, are also well within the limits of the device.
By combining it with a database of nutritional information, our proposed system can be used to help 
individuals who follow a medical diet maintain this diet while traveling.

\end{abstract}



\section{Introduction}
\label{sec:intro}
Diet plays an important role in health management~\cite{thompson2008}.
Indeed, the symptoms or risk factors of many diseases can be decreased by diet modification. In some extreme cases, for example peanut allergies, the consumption of even a minute amount of certain nutrients can be fatal. In other cases, such as diabetes or inborn errors of metabolism, the consumption of certain nutrients must be carefully monitored and limited in order to maintain an individual's health. 

In unfamiliar settings, in particular when traveling to a foreign country, maintaining a diet
can be a challenge.
Gastronomy often varies from region to region and so tourists naturally expect to be confronted with unknown dishes and ingredients. 
But while many consider sampling the local gastronomy an important part of the travel experience, people on a diet are often reluctant to embark on such journeys for fear of putting their health at risk. Diets are especially difficult to deal with when traveling to a region where a foreign language is spoken. 
Without the ability to understand menus, it is impossible to make informed food choices.
Moreover, even if the traveler is able to translate the items, the validity of the translations is usually affected by differences in culture, language, and educational levels~\cite{Resnick_Luisi_Vogel_2009}.
The importance of this problem is the topic of several books~\cite{lowell2005gluten, koeller2011gluten}.
A device capable of automatically translating menus in real-time could thus be a useful tool for people following a diet. Unfortunately, the most effective electronic translators typically rely on a remote network-connected server to obtain the translation. Connectivity issues, including the high international data plan rates, make this an unattractive solution in foreign countries.
Moreover, few automatic translators give context specific information, let alone diet specific information. As an illustration, Table \ref{tab:trans} shows the translation of a typical Spanish dish, \textit{pesca\'ito frito}, using different well-known online translators. Six out of seven translations are incorrect, and the only correct translation (Google Translate) provides little information (\textit{pesca\'ito frito} is not merely fried fish, but a special type of \textit{battered} fried fish, typically cod, which is a traditional Shabbat dish).

\begin{table*}[htb] \small \renewcommand{\arraystretch}{1.3}
\centering
\caption{Translations of the Spanish \textit{pesca\'ito frito} into English using different translators.} 
{
\begin{tabular}{|c|c|}
\hline
\textbf{Online translator} & \textbf{English output} \\
\hline
Google Translate & fried fish \\
\hline
WorldLingo, SDL FreeTranslation, Yahoo Babel Fish & pesca\'ito fried \\
\hline
PROMT Translator & pesca\'ito fried food \\
\hline
Tranexp & pesca\~A to fried \\
\hline
Intertran & fishing fried \\
\hline
\end{tabular}
}
\label{tab:trans}
\end{table*}

The problem of maintaining a diet in a foreign setting goes beyond translation: it is about interpretation, disambiguation, and communication.
For example, the text descriptions of the food items offered in a menu leave room for interpretation, even for a person fluent in the local language. This is because the name of a dish may not be descriptive, or the ingredients used to prepare the dish may vary in a particular region or even at a particular restaurant. Furthermore, certain diets involve strict preparation guidelines  (e.g., to avoid cross-contamination), and ensuring that these guidelines are followed necessitates a non-trivial dialog with the restaurant staff.

\begin{figure}[htb]
\begin{center}
\includegraphics[width=0.40\textwidth]{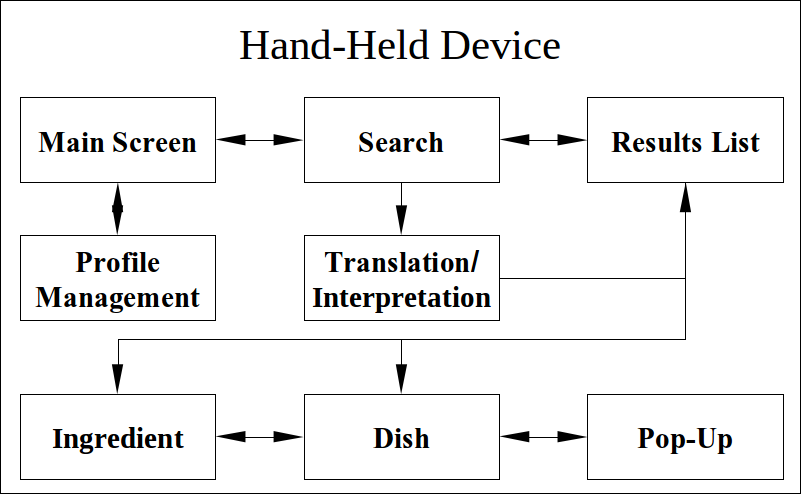}
\caption{{Block diagram of our proposed menu translation and interpretation system.}  
 }
\label{fig:restmenusbd}
\end{center}
\end{figure}

\begin{figure*}[ht]
\begin{center}
\subfloat[]{\label{fig:iphoneA}\includegraphics[width=0.18\textwidth]{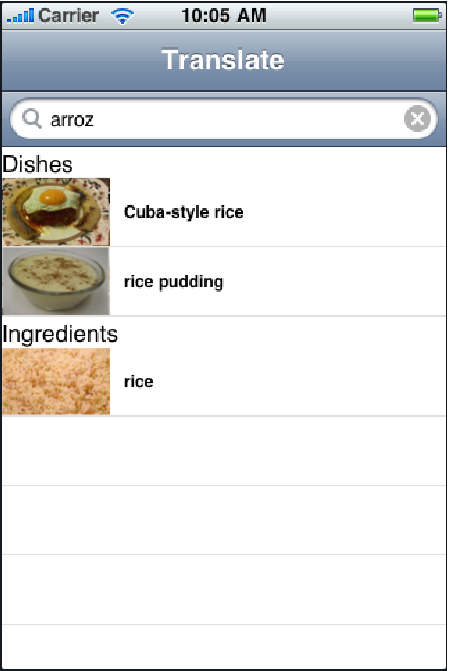}}
\hspace{1pt}
\subfloat[]{\label{fig:iphoneB}\includegraphics[width=0.18\textwidth]{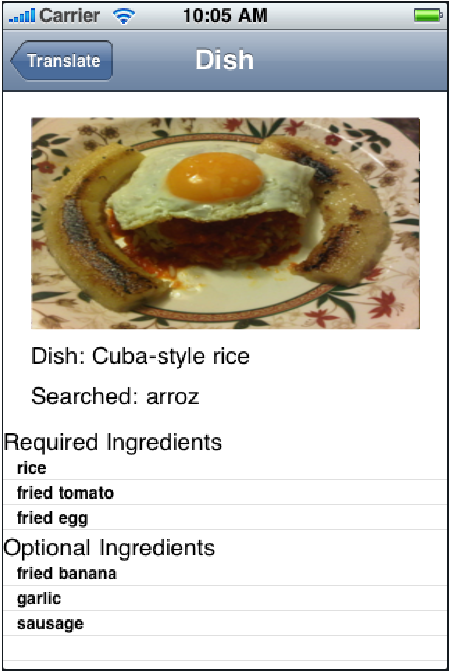}}
\hspace{1pt}
\subfloat[]{\label{fig:iphoneC}\includegraphics[width=0.18\textwidth]{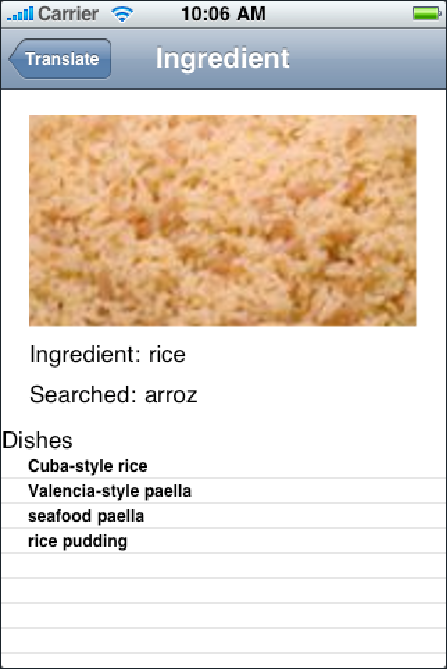}}
\caption{Proposed translation system implemented on an iPod Touch.}
\label{fig:iphoneScreen}
\end{center}
\end{figure*}

With this in mind, we propose a hand-held translation system. Our system, summarized in Figure \ref{fig:restmenusbd}, build on preliminary work previously presented in ~\cite{parra2011icme}. It is based on the use of a hand-held multimedia device such as a PDA or a mobile telephone.
Figure \ref{fig:iphoneScreen} shows screenshots of our implementation of the system in an iPod Touch Second Generation.

Real-time text translation using a hand-held device is a challenge because of the limited amount of data storage, memory (RAM), and CPU available. The use of statistical machine translation engines usually involves the use of large training data. This data consists of a pair of aligned texts, in the source and target languages, called \textit{corpora}. After training with large corpora, the resulting phrase-table (a table of statistical relationships, or weights, between the source and the target languages) can be very large, sometimes so large that it takes hours for a desktop computer to perform a translation. 
We address this challenge by using context-specific (either manually or automatically built) corpora. 
In our method, two steps we call \textit{translation standardization} and \textit{n-gram consolidation} are applied to this corpora before the training process in order to improve the statistics of the phrase-table and decrease the complexity of the search. The resulting phrase-table is orders of magnitude lighter than the ones commonly used in market applications. This also makes the translation computationally less expensive, thus decreasing battery usage.

Another challenge is translation accuracy. 
By implementing the translation standardization and the n-gram consolidation method, the number of words in the corpora is decreased, while the relationships between entries in the language pairs stay the same. 
This yields a better training, and therefore better statistics, which enhances the translation accuracy. 
The accuracy is also increased by displaying the output as a rated list of best translations, thus enabling the user decide which result better fits the context.

Translation ambiguity is another concern. 
We address this issue with the use of a browsable multimedia database that provides additional information to the user (images and lists of ingredients).


Our application can be used in a real-time network-independent fashion while efficiently producing fast and accurate results.
The computational cost and memory requirements of our methods are low enough so that the system can be implemented using many commercially available hand-held devices. To demonstrate this, we developed and implemented a prototype of this system on the iPod Touch for English speakers traveling in Spain. Our tests indicate that our translation method yields  the correct translation more often than general purpose translation engines such as Google Translate, and does so almost instantaneously. The memory requirements of the application, including the database of images, are also well within the limits of the device.

This paper is divided as follows. First, a review of existing diet management tools is presented in Section \ref{ssec:existingmethods-diet}. A summary of machine translation methods is given in Section \ref{sec:existingmethods}. The details of our proposed system are given in Section \ref{sec:proposedsys}.
We outline the implementation of our prototype in Section \ref{sec:sysimplem}. The experimental results are presented in Section \ref{sec:experiments}. Conclusions and thoughts on future improvements are given in Section \ref{sec:conclusions}.



\section{Existing Methods}
\label{sec:existingmethods}

\subsection{Diet management}
\label{ssec:existingmethods-diet}
Until recently, diets have been managed primarily using (printed) diet-specific food databases~\cite{thompson2010}.
In this traditional scenario, the individual reads the nutrition facts and ingredients printed on the food label in order to analyze the content of the food with the help of the database.
However, without access to the nutrition label (e.g., in a restaurant), the user must question the people who prepare the food in order to obtain the required information. This typically involves a dialog between the chef/food preparation staff and the individual.

With the widespread availability of smartphones and other multimedia hand-held devices, many electronic tools are now available to assist individuals who must follow a diet~\cite{BousheyTech}. For example, text messaging can be used to send reminders to diabetes patients. One can also build on the Bluetooth capabilities of such devices to remotely record blood pressure readings. With the higher resolution pictures, improved memory capacity, and faster processors in recent versions of these smart devices, it may even be possible to automatically identify and measure the food consumed from "before and after" pictures of a plate~\cite{zhu2010,six2010}.

There are also smartphone applications targeting individuals who must record their intake of certain nutrients such as calories (e.g., myFitnessPal) or amino acids (e.g., DietWell for PKU). However, when traveling to a foreign region, there is an idiomatic barrier that has to be surpassed before being able to use any of the tools mentioned above. Traditional solutions include pocket dictionaries or translation cards~\cite{koeller2010multi}. However, pocket dictionaries are not specific enough for certain medical diets, and only translate single words; translation cards include terms specific to the user's medical conditions, but they are just a tool to communicate with foreign individuals, and they do not offer translations for restaurant menu items.

\subsection{Automatic Translation}
\label{ssec:existingmethods-translation}

Machine translation (MT) methods fall into two main categories: Rule-Based MT (RBMT) and Statistics-based MT (SMT). RBMT provides a text translation based on the grammatical, morphological and syntactical rules of the languages in question. The advantage of this method is that it deals with grammar rules and lexicon, so almost any morphosyntax variation of an input can be handled. The disadvantage is that an extensive knowledge in both the source language and the target language is required in order to build the rules, and a lot of human effort has to be invested in creating the rule-based dictionary. In contrast, SMT systems only need a parallel corpora in the source language and the target language. This corpora is then trained and a phrase-table is created, indexing the probability weights of the most relevant n-grams in the corpora. Along with the phrase-table, a language model and a reordering table are also created in order to complement the decoding process. The disadvantage is that it usually relies on large corpora (billions of words) to provide a good quality translation. This corpora has to be manually built. Fortunately, there are many freely available corpora.

The results of the \textit{2009 Workshop on Statistical Machine Translation} for the open source systems indicate that Apertium and Moses~\cite{2009Workshop} are among the best representatives of RBMT and SMT, respectively. We evaluated them separately. \\

\subsubsection{RBMT: Apertium}
\label{ssec:existingmethods-apertium}
The open source RBMT system Apertium is a shallow-transfer engine that uses hidden Markov models (HMMs)~\cite{1164070}~\cite{1583227} for part-of-speech (POS) tagging and finite state transducers~\cite{4220679} for lexical transformations. It also works with Constraint Grammar~\cite{991176} taggers for some language pairs. The project's original purpose was the translation between closely related languages (e.g., Spanish-Catalan, Basque-Spanish, Galician-Spanish, Danish-Swedish, Czech-Slovak) but it has been expanded to support more common languages (e.g., English, French, Portuguese).

\indent
One advantage of Aperium is that, being rule-based, it does not need multiple entries for different forms of a word. To add a word (and all its forms) one just needs to add one entry containing the prefix of the word that is common to all inflected forms, and the inflection paradigm of the word.
%
%
One disadvantage is that each entry has to be typed in three different files and in three different ways (Spanish monolingual dictionary, bilingual dictionary and English monolingual dictionary). Also, the rule-based tables do not contain accurate food related information, hence a lot of the food specific entries would have to be manually entered. For example, the Apertium Spanish-English dictionaries contain only about 5,000 of the 7,000 food-related items we would need to accurately translate Spanish dishes. These 2,000 entries would have to be manually added. Another disadvantage is that linguistic rules do not generalize to other languages, so manual development is necessary for every new language pair. Taking everything into consideration we decided not to use Apertium as the basis of our system. \\

\subsubsection{SMT: Moses}
\label{ssec:existingmethods-moses}
The main features of the Moses engine are confusion network coding~\cite{Mangu00findingconsensus} and word lattices~\cite{1613792}, allowing the translation of ambiguous inputs. It also uses factored translation models~\cite{koehn-hoang2007}, making it able to add part-of-speech tags or lemma information to the phrase-table. Moses uses external tools for word alignment (GIZA++~\cite{Och03asystematic}) and language modeling (SRILM~\cite{KoehnHB}) as part of the training process. It also uses a beam-search heuristic algorithm to find the highest probability translation among the exponential number of choices (roughly similar to ~\cite{Tillmann03}). As stated earlier, the main advantage of the SMT method is that it only needs two files (or corpus): one in the source language and one in the target language. The different training tools prepare and process the data so to obtain a phrase-table with all the probabilities as well as other information needed for translation~\cite{zens-ney2007}.
%

Generally, the training data has to be large enough for effective training. However, when working in specific domains such as restaurant menus the corpora can be reduced, which in turn reduces the size and translation time of the resulting translation application. Further improvements can be obtained following the method we propose in Section \ref{sec:proposedsys}.


\section{Proposed System}
\label{sec:proposedsys}
Before leaving for a foreign country, the user must download a region and language specific configuration and database. From then on, the system operates without a network connection.
When considering a given dish or food item on a foreign menu, the user then enters the text of the menu describing this item (in the local language). This is done using hardware or virtual input on the device (e.g., keyboard or handwriting recognition, etc.). The text entered is then translated by the device. The translation is conveyed to the user using a multimedia display including text and still images. When appropriate, disambiguation of the translation results is handled through a device enabled bilingual dialog, where information or instructions to be transmitted to the waiter are suggested by the device.

\subsection{Menu-specific corpora}
\label{ssec:proposedsys-menuSpecificCorpora}
SMT engines typically require large training data in order to be accurate. Unfortunately, the larger the training set, the larger the amount of memory required (e.g., phrase-tables, language models, etc.). For example, the \textit{WMT08 News Commentary} has a corpora of 15 MB, which produces 708 MB of data to be used during the decoding process. Obviously, this is too large for most hand-held devices. However, in context-specific applications a more modest size corpora can be used. In our case, our corpora focuses on restaurant menu items, for which the vocabulary is rather limited.

Observe that menu items are described using short sentences (typically less than 5 words). Thus, corpora focused on menus contain fewer words per line than other, non-context specific corpora. For example, sentences in the \textit{WMT08 News Commentary} have, on average, 21.6 words. Since shorter sentences yield lighter phrase-tables, we can expect a small phrase table size when using a menu-focused corpora. In the following, we will describe some pre-processing steps to be applied to the corpora so to further decrease the size of the phrase-table while maintaining, or even increasing, accuracy.

\subsection{Translation standardization and corpora division}
\label{ssec:proposedsys-ptoptimization}
The main idea behind SMT is to find the most likely translation $\hat{s}$ in the source language given a phrase $t$ in the target language. The probability that a phrase $s$ is the translation of the phrase $t$, $p(s|t)$, is (by Bayes Theorem) $p(s|t) = \frac{p(s)p(t|s)}{p(t)}$. Since the denominator $p(t)$ is independent of $s$, finding the most likely phrase $\hat{s}$ is the same as finding the phrase $s$ that maximizes the product $p(s)p(t|s)$. We thus obtain the Fundamental Equation of Machine Translation~\cite{Brown93}, which allows us to obtain the most likely translation $\hat{s}$, given by $\hat{s} = \arg\max\limits_s [p(s)p(t|s)]$. In order to find $\hat{s}$, the system has to perform an exhaustive search in the phrase-table by going through all the strings $s$ in the source language. This is one reason why the corpora has to be built carefully so that, after training, we obtain a small phrase-table.

An n-gram linguistic model is used to approximate the language model ~\cite{srilm}. In this model, the probability of a word in a sequence depends on the previous words in the sequence $p(x_i|x_{i-1},x_{i-2},...,x_{i-n})$.

%
%
%
%
%
%
%

By using n-grams with $n>1$ the model accounts for word reordering. 
The higher $n$, the higher the number of reorderings accounted for. 
However, the higher $n$, the higher also the number of words ($n-1$) the probability of the last word is conditioned on. This increases the difficulty of estimating $p(x_i|x_{i-1},...,x_{i-n})$, as for large $n$ the condition $x_{i-1},...,x_{i-n}$ has little chance of occurring in the corpora.

When training using a SMT engine like Moses, the resulting phrase-table contains the n-grams that appear most often in the corpora. Each i-gram ($i \in [1,n]$) in the source language is associated to one or more j-grams ($j \in [1,n]$) in the target language, and each combination has an associated probability weight in the phrase-table. These associations and probabilities are obtained automatically by analyzing the corpora. The more possible translations an n-gram has, the more entries are created in the phrase-table, and the more disperse the probabilities are. This leads to three important problems: first, the phrase-table becomes larger as the number of possible translations increases, so large that it may not fit in the memory of a hand-held device. Second, all possible translations must be considered when translating, so this increases the computational search time needed to perform a translation. Third, since the same n-gram in the source language has more possible translations, the correct translation for a specific input is less likely to be considered by the decoder.

Observe that sometimes an n-gram in the source language does not literally correspond to any n-gram in the target language. For example, consider the pair \textit{arroz a la cubana} $\leftrightarrow$ \textit{rice with fried eggs and banana fritters}. In this instance, the Spanish 1-gram \textit{arroz} corresponds to the English 1-gram \textit{rice}, and the Spanish 3-gram \textit{a la cubana} corresponds to the English 6-gram \textit{with fried eggs and banana fritters}. Thus, an entry in the phrase table will consist of the pair \textit{a la cubana} $\leftrightarrow$ \textit{with fried eggs and banana fritters}. However, this n-gram association is only valid for this particular dish. Therefore, this pair will have a low probability weight. Since \textit{a la cubana} has different meanings for different dishes (for example	\textit{arroz a la cubana} $\leftrightarrow$ \textit{rice with fried eggs and banana fritters}, or \textit{garbanzos a la cubana} $\leftrightarrow$ \textit{chickpeas with chorizo, ham, chicken and vegetables}) multiple entries would be created in the phrase-table for the Spanish 3-gram \textit{a la cubana}, each one having its own probability weight. However, one can also translate \textit{a la cubana} as \textit{Cuba-style}.


\noindent If this is done for every dish containing the expression \textit{a la cubana}, the entry \textit{a la cubana} will appear only once in the phrase-table, with \textit{Cuba-style} as a translation, and the translation will be what we call \textit{standardized}. The disambiguation for each particular dish can later be done by pulling information from the database of ingredients and images.

Standardized translations help address the three aforementioned issues. First, when an n-gram is standardized it only has one possible translation, hence an input containing a standardized n-gram will always produce the same translation. Second, since there is only one possible translation for the standardized n-gram, only one entry is going to be created in the trained phrase-table for this n-gram. Therefore, the computational search time is reduced. Third, by having only one entry in the phrase-table for a standardized n-gram, the size of the phrase-table is smaller than if we had multiple possible translations for the same n-gram.

There is a trade-off between the number of standardized translations in the corpora and the database resources required to disambiguate them. The higher the number of standardized translations, the lower the computational search time, the more accurate the statistics are, and the smaller the trained phrase-table is. However, the higher the number of standardized translations, the more disambiguations are required, but the trade-off is worth it. Thus we propose to either favor standardized translation over non-standardized ones when building the corpora, or replaced non-standardized translations by standardized ones whenever possible.

We also propose to divide our corpora into two in order to optimize the training so the phrase-table contains fewer and more accurate entries. More specifically, we only used standardized translations as part of the training data; non-standardized translations are directly moved into one-to-one phrase-tables put together with the phrase-table obtained by training. Thus, we obtain two different bilingual data sets: \\

\begin{enumerate}
	\item Set 1: training data. When the translation is either fully or partly word to word (e.g., \textit{tortilla de patatas} $\leftrightarrow$ \textit{potato omelette}) (i.e., the translation is standardized), the phrase pair is placed in the training data, which will influence the probability weights of the trained phrase-table. Such phrase pairs are grammatically correlated, and the relationship between their words can be extended to other cases (e.g., \textit{tortilla de espinacas} $\leftrightarrow$ \textit{spinach omelette}). As a result, the phrase-table contains fewer possible translations for the same n-gram (e.g., \textit{tortilla} $\leftrightarrow$ \textit{omelette}), thus increasing its probability weight. \\
	
	\item Set 2: one-to-one phrase-tables. When the translation is not standardized (i.e., when it is not literal, such as \textit{caf\'e cortado} $\leftrightarrow$ \textit{espresso with milk}), the phrase pair is included in a one-to-one non-trained phrase-table. The tables, which are analogous to the trained phrase-table, are divided by topics and simply added to the phrase-table obtained by training. Its entries are not n-grams, but entire phrases. However, the decoder does not distinguish between n-grams and phrases. The probability weights of each phrase pair are set to 1.0. Note that this is a probability weigh, not an absolute probability. This generally gives the phrase pair more weight than it would get if they were included in the training data set, so they are more likely to be considered for translation. \\
\end{enumerate}

Note that it is possible for a phrase to appear both as an n-gram in the training data and as part of some n-grams in a one-to-one phrase-table. To understand how this is handled by the decoder, consider the following two possible translations for the Spanish food-related items containing the word \textit{cortado}: \textit{yogurt cortado} $\leftrightarrow$ \textit{sour yogurt}, and \textit{caf\'e cortado} $\leftrightarrow$ \textit{espresso with milk}.
The first translation, \textit{sour}, is the most common one. But there are cases where this translation is clearly not appropriate (e.g., in the second case). If both cases are included in the training data set, and assuming that \textit{cortado} $\leftrightarrow$ \textit{sour} appears more frequently than \textit{cortado} $\leftrightarrow$ \textit{espresso with milk} in our corpora, the probability weights in the phrase-table are going to be weighted strongly in favor of \textit{sour}. That is, $p_{\textit{cortado} \rightarrow \textit{sour}} = w_1$, and $p_{\textit{caf\'e cortado} \rightarrow \textit{espresso with milk}} = w_2$, with $w_1 \gg w_2$. Therefore \textit{caf\'e cortado} will probably be translated incorrectly as \textit{sour coffee}, instead of \textit{espresso with milk}. Instead, if \textit{caf\'e cortado} $\leftrightarrow$ \textit{espresso with milk} is removed from the training data set and included in a one-to-one phrase table, the probabilities do not interfere with each other, hence $p_{\textit{cortado} \rightarrow \textit{sour}} = 1.0$ in trained phrase-table, and $p_{\textit{caf\'e cortado} \rightarrow \textit{espresso with milk}} = 1.0$ in one-to-one phrase-table. Therefore, \textit{caf\'e cortado} is likely to be translated as \textit{espresso with milk}. It is not guaranteed that the correct translation would be obtained. Indeed, the probabilities set in both the trained phrase-table and the one-to-one phrase-tables are probability weights, not an absolute probabilities. These weights always have an influence on the decoding process, as well as the language model, the distortion model, and the word penalty.
%
%
%

Once the training process is complete, the resulting phrase-table contains n-gram pairs (source and target languages) with automatically estimated weights. The translation of a phrase is then the translation with smaller cost among all the generated translation hypotheses. The most likely translation may not be the one given by the one-to-one phrase table, but it should be close to it. We thus configure Moses to provide a \textit{best-list} of results (multiple outputs), and the user can use his/her best judgment to determine the most appropriate using disambiguation information (ingredients list, dishes/ingredients images, dialogs with a native speaker, etc.).

\subsection{n-gram consolidation}
\label{ssec:proposedsys-ngram}
Increasing the corpora size with new dishes is a way to increase the accuracy of the translation. But, because of the additional time and computational cost related to the use of the larger phrase tables resulting from training with more data, we propose another way to obtain better accuracy while maintaining or even reducing the corpora. The idea is to match the number of n-grams in both the source language phrase and the target language phrase in order to increase the probability of success. For instance, the Spanish-English pair \textit{crema a la menta} $\leftrightarrow$ \textit{mint cream} has four words for Spanish and two words for English (4 $\leftrightarrow$ 2), and the relationship is \textit{crema} $\leftrightarrow$ \textit{cream} (1 $\leftrightarrow$ 1), and \textit{a la menta} $\leftrightarrow$ \textit{mint} (3 $\leftrightarrow$ 1).
%
%

Table \ref{tab:posindex} shows some examples of translations and their corresponding position indices, indicating word relationships. Moses takes all the n-grams (where $n$ is set during the training process to create the language model through SRILM) on the corpora and creates a table of probabilities, giving more weight to the most frequent combinations. However, as seen in Table \ref{tab:posindex}, there are lots of different dishes with the string \textit{...a la...} and different position indices.

\begin{table}[htb] \small \renewcommand{\arraystretch}{1.3}
\centering
\caption{Examples of position indices.}
\begin{tabular}{|c|c|c|}
\hline
\textbf{Spanish} & \textbf{English} & \textbf{position indices}  \\
\hline
\textit{arroz a la cubana} & \textit{Cuba-style rice} & 0=1, 1=0, 2=0, 3=0 \\
\hline
\textit{crema a la menta} & \textit{mint cream} & 0=1, 1=0, 2=0, 3=0 \\
\hline
\textit{pato a la naranja} & \textit{duck \`a l'orange} & 0=0, 1=1, 2=2, 3=3 \\
\hline
\textit{cordero a la miel} & \textit{lamb with honey} & 0=0, 1=1, 2=1, 3=2 \\
\hline
\end{tabular}
\label{tab:posindex}
\end{table}
\normalsize

There is not a defined structure for these cases, hence there is not a default translation for the Spanish phrase \textit{...a la...}. Even if we use corpora with millions of words in them, we still would not have a default translation for this phrase.
Moreover, even if enlarging the corpora to millions of words would give more accurate probabilities, it would make the resulting phrase-table size increase to gigabytes.

We propose the reduction of n-grams with large $n$ by merging multiple words in a phrase, thus balancing the indices in both the source language and the target language sides. For example, instead of trying to find a general translation for the general structure \textit{...a la...} (which does not have a translation in English), it is better to focus on subsequences of words for which there is a defined translation. For example, by merging the combination \textit{a la cubana} down to one word (concatenated with the string $\&$) on the source language side of the corpora (i.e., \textit{arroz a\&la\&cubana} $\leftrightarrow$ \textit{Cuba-style rice}) a possible 4-gram (16 possible matches) on the source language phrase is reduced to a simple balanced (0=1, 1=0) 2-gram (4 possible matches). By doing this, we achieve two important goals. First, we are weighting the distribution of n-grams towards low values of \textit{n}. Second, we decrease the number of words in the corpus. We call this method n-gram consolidation.

After the merging step, the probabilities are better distributed and more translations are correct. See Tables \ref{tab:newprob} and \ref{tab:defo} in Section \ref{ssec:experiments-ngramanalysis} for detailed results and more examples. The words to be merged have to be chosen very carefully to obtain good statistics. For example, we make sure that every time the phrase \textit{...a la...} appears in the Spanish corpus without being merged into \textit{...a\&la...} it is because its translation is \textit{...to the...}. This makes the trained phrase-table assign the pair \textit{...a la...} $\leftrightarrow$ \textit{...to the...} with 1.0 probability weight. In some cases there is more than one translation associated with the same input, such as for the Spanish \textit{...al...}. Again, we make sure that it is only translated to either \textit{...with...} or \textit{...au...}. Also note that the n-gram consolidation is applied to the target language corpus.

The n-gram consolidation has another advantage: by consolidating, the distribution of n-grams is moved towards smaller $n$ (depending on the language, often $n=1,2,3$). This gives more weight to the n-grams with smaller $n$. Therefore, the translation can be accurate with much fewer training pairs containing \textit{...a la...}.

The target part of the corpus has to be modified in accordance with the modifications induced by the n-gram consolidation in the source language. In order to do so, we need to specify in what cases we perform the merging. That is, whenever two phrases are equivalent in both the source and target languages, but they have different number of words, the longer phrase is consolidated to match the length of the shorter. These cases are specified in a list manually built by a person expert in both languages. This list, which contains all the n-gram consolidations that have been used in the training data set of the corpora, is built so the specific cases are treated first (e.g., \textit{...a la cubana...} $\leftrightarrow$ \textit{...a\&la\&cubana...}), followed by the general cases (e.g., \textit{...a la...} $\leftrightarrow$ \textit{...a\&la...}). Note that the n-gram consolidation can be used in entries in both the source and the target languages. 

\subsection{Automatic n-gram consolidation}
\label{ssec:proposedsys-ngram-auto}
The previously described manual n-gram consolidation method, significantly reduces the size of the phrase-table used by the decoder. However, it is desirable to automate the process, since it would make it simpler to expand the current corpora or to create new corpora for different language pairs. Below we propose such an automatic n-gram consolidation method.

The goal of automatic n-gram consolidation is to emulate the consolidation decisions made by a human expert. We have attempted to do this consolidation in a language independent fashion. That is, we tried to create a consolidation method that would work without prior knowledge of the language or knowledge of the types of phrases to be consolidated.

Notice first that not all phrases in a corpus should be consolidated. Only the phrases whose translation has fewer words than the source language input could potentially benefit from consolidation. A decision regarding which phrases could potentially be consolidated is made using a comparison of the length of each phrase in the source language corpus with its translation in the target language corpus. More specifically, if a phrase in the target language corpus has fewer words than the phrase in the source language corpus the source language phrase is marked for potential consolidation.

Next, among the phrases marked for consolidation, a search for common n-grams (subphrases) is made. Each consecutive word in each n-gram in the set of marked phrases in the source language is compared with the source language phases (n-grams) in the marked training set. This particular computation tends to be the most computationally complex portion of the whole process at $O(k^2n)$ where $k$ is the number of phrases in the marked training set and $n$ is the number of words in each phrase. After we have found all common phrases in the marked training set, we replace the common phrases with their respected consolidated n-gram.

Note that the n-grams chosen for consolidation may not necessarily correspond to the n-grams chosen by the manual consolidation, though it is likely that there will be an intersection between the n-grams found automatically and manually. But both methods achieve the goal of moving the distribution of n-grams towards smaller $n$.

\subsection{Application to diet management}
\label{ssec:proposedsys-multimedia}
Multimedia in diet management offers great advantages over traditional methods. For example, images and ingredients lists can be used in our system to provide diet-related guidance in a foreign country. Indeed images can disambiguate the result of a translation while ingredient lists can warn of the presence of certain nutrients. Moreover, the touch screen can be used to enable a bilingual dialog between two monolingual individuals.

\begin{figure*}[ht]
\begin{center}
\subfloat[]{\label{fig:iphoneAdoc}\includegraphics[width=0.18\textwidth]{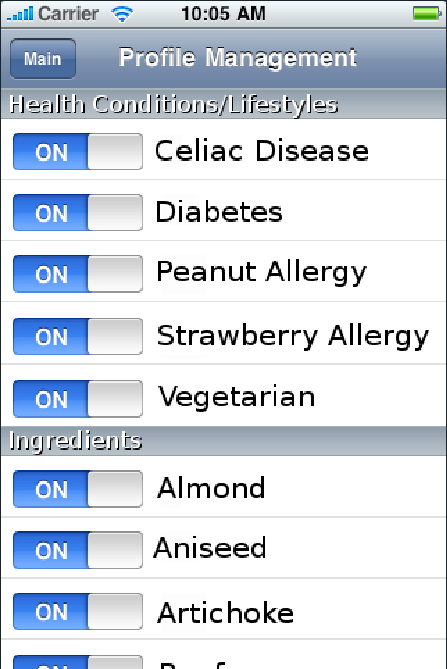}}
\hspace{1pt}
\subfloat[]{\label{fig:iphoneBdoc}\includegraphics[width=0.18\textwidth]{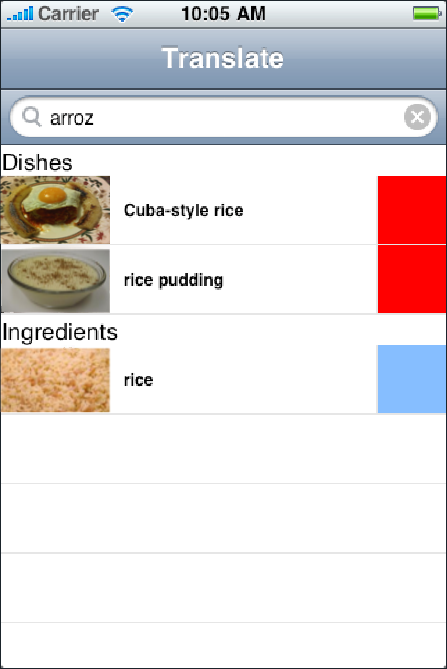}}
\hspace{1pt}
\subfloat[]{\label{fig:iphoneCdoc}\includegraphics[width=0.18\textwidth]{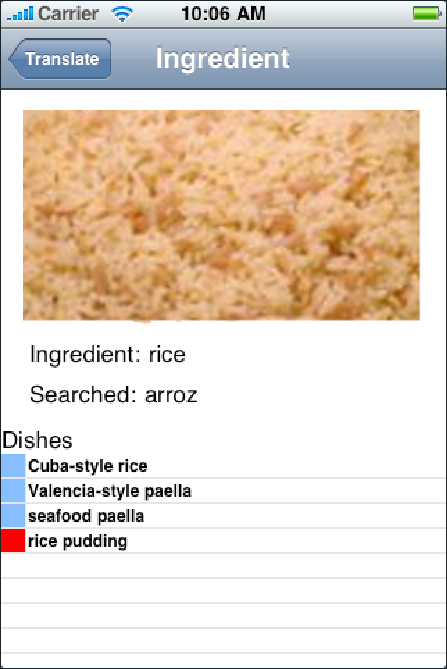}}
\hspace{1pt}
\subfloat[]{\label{fig:iphoneDdoc}\includegraphics[width=0.18\textwidth]{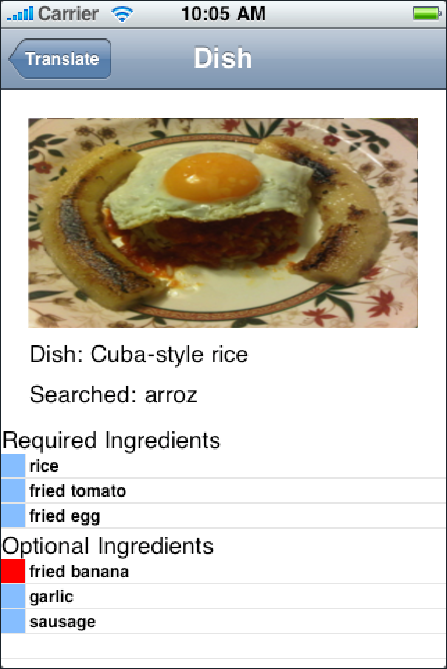}}
\hspace{1pt}
\subfloat[]{\label{fig:iphoneEdoc}\includegraphics[width=0.18\textwidth]{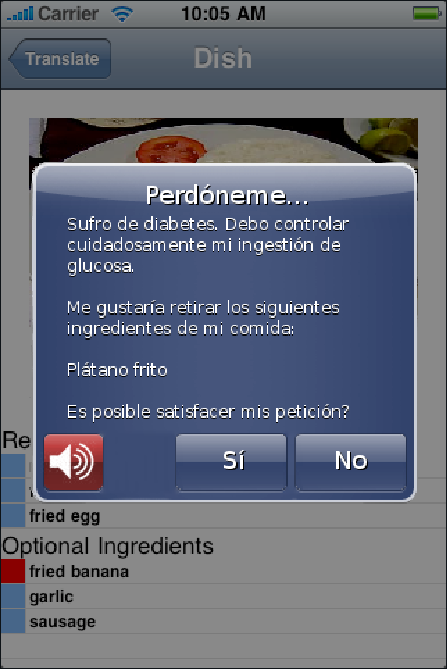}}
\caption{Proposed extension of our system to diet management (images modified to illustrate profile management, ingredient flagging, and bilingual dialog).}
\label{fig:iphoneScreenDoc}
\end{center}
\end{figure*}

Figure \ref{fig:iphoneScreenDoc} illustrates an extension of our translation system on an iPod Touch device including additional multimedia elements to facilitate diet management.
Before using the application, the user would have to create a diet profile. On the device, a list of conditions/allergies/lifestyles would be shown through the GUI, as well as a list of all the ingredients in the database (Figure \ref{fig:iphoneAdoc}). Upon selecting either a diet profile and/or some ingredients, the database would be flagged to reflect the user's diet (see Section \ref{sssec:sysimplem-appdev-db}). Thereafter, if a selected dish may contain one of the ingredients in the user's profile, it would appear flagged Figures \ref{fig:iphoneBdoc} to \ref{fig:iphoneDdoc}), and the user could decide whether to choose another dish, request to remove the flagged ingredients, or ask for clarification. Both the ingredient removal request and the clarification could be accomplished by clicking on a suggested question dialog on the screen; the question would then be displayed in the local language along with possible answers to be selected by the local staff (Figure \ref{fig:iphoneEdoc}). Once selected, the answer of the staff would be translated back to the user.

Note that the current implementation of our system includes neither the profile management nor the dish and ingredient flagging, so Figures \ref{fig:iphoneAdoc} to \ref{fig:iphoneEdoc} have been modified to show the proposed scenario.

\section{Translation System Implementation}
\label{sec:sysimplem}

We implemented a prototype of our system on a Second Generation iPod Touch and with the Spanish (Spain) to English (USA) language pair. Our proof of concept implements our proposed translation module for Spanish to English translation and a small database used for translation disambiguation. The database comprises some ingredient lists and dish/ingredients images. We chose to implement the proposed translation module and the disambiguation database in order to determine the feasibility of our system. In Section \ref{sec:experiments}, we analyze the speed, accuracy and size of our translation module and extrapolate the multimedia size requirements for a full size disambiguation database. We did not implement the diet management portion of our proposed system as this would be a straightforward addition to the system.

The input of the dish or ingredient is through the iPod virtual keyboard. However, this could also have been done through the use of a hardware keyboard or Optical Character Recognition (OCR), although OCR would require additional computations.


\subsection{Corpora Acquisition and SMT training}
\label{ssec:sysimplem-corpacq}
In order to obtain the phrase-tables required to perform translations, we created our own Spanish-English corpora. We obtained a bilingual list (corpora) of dishes by searching free online sources. We also read several Spanish restaurant menus and manually translated their listed dishes into English. Our corpora\footnote{Corpora available online: https://redpill.ecn.purdue.edu/$\sim$reddo/resources} contains a total of 5,854 entries, with 2.83 words per entry on average in Spanish, and 2.07 words per entry on average in English. Single ingredients comprise 2,216 entries, and the remaining 3,638 entries are composite dishes. Note that, due to the nature of the SMT paradigm, it is not necessary to include all possible combinations of dishes and ingredients. However, it is important that all the declensions of each word (i.e., grammatical case, number, and gender) are included in the corpora. We manually divided this initial corpora into two sets. One set contains the standardized phrase pairs, which will form the corpora (training data) used for training. The other set contains the remaining phrase-pairs, formatted as one-to-one phrase-tables, and divided by topics. We trained Moses using the standardized phrase pairs and the language model in order to obtain the trained phrase-table. The one-to-one phrase tables were then manually added to the trained phrase-tables.

\subsection{Prototype development}
\label{ssec:sysimplem-appdev}
To implement the system on the hand-held device we first designed a relational database. The database is used to store images and ingredient lists along with relational information about Spanish dishes and their ingredients. Second, we created a semantic model and used it to represent the data during runtime. Third, we developed a parser to automate the population of the menu database with images and relational data. Finally, we developed a Graphical User Interface (GUI). The user's interaction with the GUI is bidirectional, since all the data is connected and the user can switch dishes and ingredients, and access additional information.\\


\subsubsection{Moses incorporation}
We made several modifications to the Moses source code in order to incorporate it into the iPod touch. 
We used Moses to memory-map the language model (LM) and the phrase-tables. 
This feature allows the system to process and read the queries from disk on-demand instead of being loaded entirely into memory. 
Therefore, only the phrase-table pairs required to translate the input are loaded into memory.

\subsubsection{Database design}
\label{sssec:sysimplem-appdev-db}
We normalized the relational database. It consists of six tables: ingredients, dishes, images, relationships between ingredients and dishes, dishes and images, and ingredients and images. The engine used is an SQLite embedded relational database management system because the Application Programming Interface (API) is built into the iPod.


\begin{figure}[htb]
\begin{center}
\includegraphics[width=0.48\textwidth]{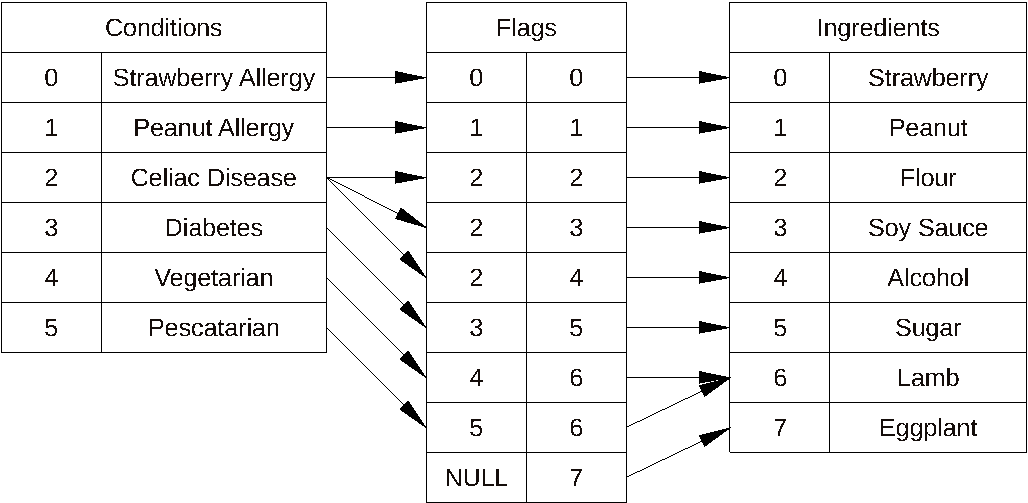}
\caption{Schema for the ingredients and conditions relationship.}
\label{fig:flagging}
\end{center}
\end{figure}

\subsubsection{Semantic model}
\label{sssec:sysimplem-appdev-class}
Similarly to the database, it is important to capture the real-world representation and relationships of the data in order to represent it in an effective, intuitive, and meaningful way in software. We represent dishes, ingredients and images, as well as some of the utility and user interface classes engineered for the system, as classes. Each dish and ingredient inherits functionality from two interfaces (protocols in Objective-C): \textit{UIObject} and \textit{DatabaseObject}. The methods inherited by the \textit{DatabaseObject} interface enforce the required database access methods (insert, update, select and delete) while the \textit{UIObject} interface allows each object to create a context specific view of itself on the screen.\\

\subsubsection{Data parsing}
\label{sssec:sysimplem-appdev-parsing}
In an effort to automate the process of populating the database with data on dishes, images and ingredients, a very simple Domain Specific Language (DSL) has been written to represent the database in text form. The DSL uses a series of symbols to represent the relationship between ingredients, images, and dishes in the database, as in the following example: \\


\begin{center}
	\fbox{
		\parbox{0.20\linewidth}{
			\textit{\#bread with tomato \\
			-bread \\
			=toasted bread \\
			-tomato \\
			-olive oil \\
			\$oil \\
			-salt \\
			-+garlic}
		}
	}
\end{center}

\vspace{10pt}

\begin{itemize}
\item \# Represents the name of a dish. The following lines until the next ``\#'' represent ingredients and images, many with special relationships.
\item - Represents an ingredient used in the making of a dish.
\item = Always follows a ``-''. It represents an ingredient which can be substituted for the previous ingredient.
\item \$ Can follow either a ``\#'' or a ``-''. It represents the name of an image. If this token does not appear, the name of the dish or the ingredient is assumed
to be the name of the image.
\item -+ Represents an optional ingredient in the making of this dish.\\
\end{itemize}

The DSL is first tokenized by dishes. Then, each dish is parsed ingredient by ingredient. Each ingredient contains any substitutions and images associated with it. The images in memory are loaded from image files and added as binary data to the database. The fact that an ingredient is optional and can be substituted is also noted in the database. \\

\subsubsection{GUI}
\label{sssec:sysimplem-appdev-gui}
The GUI includes functionality to enter a Spanish word or phrase into the translator, and get a list of matching results from the database (Figure \ref{fig:iphoneA}). When a dish is selected, it show a picture along with all required and optional ingredients, separated into their respective categories (Figure \ref{fig:iphoneB}). If an ingredient is selected, it shows a picture along with a list of dishes in which the ingredient can be found  (Figure \ref{fig:iphoneC}).

%
%



Blue and red rectangles could be used to indicate allowed/disallowed ingredients by the user, respectively, during the profile configuration step.
Dishes could also be tagged, depending on whether its ingredients are suitable or not for the user's diet.

While not implemented in the prototype, the database would also include a list of medical conditions or lifestyles, and relationships between the conditions/lifestyles and the ingredients that affect the condition or lifestyle. When a user wishes to add a particular ingredient for a personalized diet, a relationship would be added to the flagging table - shown by the (NULL, 7) entry in Figure \ref{fig:flagging}.




\section{Experimental Results}
\label{sec:experiments}
In this section, we analyze the speed and accuracy of the translation and interpretation modules of our system, and we report on the size requirements of the combined translation module and disambiguation database.

We tested our translation and interpretation modules on a desktop computer (AMD Turion 64 X2 Dual-Core 4GB RAM and 2.20 GHz clock rate), a Second Generation iPod Touch (ARM11 128 MB DRAM and 533 MHz clock rate), and a LG Google Nexus 5 Android smartphone (Qualcomm Snapdragon 800, 3 GB RAM and 2.26 GHz clock rate). 
We used the desktop computer to evaluate the accuracy of our translation module and the speed gain resulting from the use of our n-gram consolidated and standardized training data. 
We used the iPod Touch to evaluate the translation time and the memory requirements of an implementation on commercially available portable devices.
We used the Nexus 5 to evaluate the accuracy of both our translation and interpretation modules through a controlled user study.

Three versions of our translation module are analyzed in this section. The first version, v1, uses Moses and our corpora (described in Section \ref{ssec:sysimplem-corpacq}). The second version, v2, uses Moses along with our corpora manually standardized and manually n-gram consolidated. The third version, v3, replaces the manual n-gram consolidation with the automatic n-gram consolidation method. 
Both v2 and v3 include Moses' \textit{k-best-list} feature which outputs a list of \textit{k} possible translations sorted by their weight. In our experiments, $k$ was set to five, and the maximum length of the n-grams was set to three.

\subsection{Translation standardization and n-gram consolidation}
\label{ssec:experiments-ngramanalysis}
To analyze the effect of the translation standardization and the n-gram consolidation, we first analyzed the n-gram distribution in the training data of v2 and v3 respectively. Tables \ref{tab:v1vsv2} and \ref{tab:v1vsv3} show the number of n-gram, lines, and words in the training corpus (source language) for each version. Our results show that both manual and automatic n-gram consolidation decrease the total number of words in the training data. Moreover, in both versions the distribution of n-grams is moved towards n-grams of lower size. Although the n-grams chosen for consolidation may not necessarily correspond to the n-grams chosen by the manual consolidation, there is a large intersection between the n-grams found automatically and manually.

\begin{table}[htb] \small \renewcommand{\arraystretch}{1.3}
\centering
\caption[]{n-grams and word counts for training corpus versions 1 and 2.}
\begin{tabular}{c|c|c|c|c|}
\hline
\multicolumn{1}{|c|}{\textbf{Data set}} & \textbf{1-gram} & \textbf{2-grams} & \textbf{3-grams} & \textbf{Words} \\
\hline
\multicolumn{1}{|c|}{v1} & 2,216 & 6,452 & 1,971 & 15,182 \\
\hline
\multicolumn{1}{|c|}{v2} & 2,695 & 8,324 & 781 & 11,067 \\
\hline
 & \textbf{+21.61\%} & \textbf{+29.01\%} & \textbf{-60.37\%} & \textbf{-27.10\%} \\
\cline{2-5}
\end{tabular}
\label{tab:v1vsv2}
\end{table}

\begin{table}[htb] \small \renewcommand{\arraystretch}{1.3}
\centering
\caption[]{n-grams and word counts for training corpus versions 1 and 3.}
\begin{tabular}{c|c|c|c|c|}
\hline
\multicolumn{1}{|c|}{\textbf{Data set}} & \textbf{1-gram} & \textbf{2-grams} & \textbf{3-grams} & \textbf{Words} \\
\hline
\multicolumn{1}{|c|}{v1} & 2,216 & 6,452 & 1,971 & 15,182 \\
\hline
\multicolumn{1}{|c|}{v3} & 2,806 & 8,364 & 1,014 & 11,827 \\
\hline
 & \textbf{+26.62\%} & \textbf{+29.63\%} & \textbf{-48.55\%} & \textbf{-22.10\%} \\
\cline{2-5}
\end{tabular}
\label{tab:v1vsv3}
\end{table}

Both the translation standardization and the n-gram consolidation improve the statistics of the trained phrase-table. Table \ref{tab:newprob} shows comparisons of the probability weights obtained with v1 and v2 for the n-gram \textit{...a la...}. Notice that having standardized \textit{...a la...} in v2 creates fewer and more accurate translation pairs in the trained phrase-table. Table \ref{tab:defo} shows the probability weights obtained with all three versions for some more n-gram pairs.


\begin{table}[htb] \small \renewcommand{\arraystretch}{1.3}
\centering
\caption{Probability weights obtained by training for the standardized Spanish phrase \textit{...a la...}.}
\begin{tabular}{|c|c|}
\hline
\textbf{Phrase in v1} & \textbf{Probability}  \\
\hline
\textit{...a la...} $\leftrightarrow$ \textit{...in...} & 0.33 \\
\hline
\textit{...a la...} $\leftrightarrow$ \textit{...served with ham peppers...} & 0.17 \\
\hline
\textit{...a la...} $\leftrightarrow$ \textit{...served with ham peppers and...} & 0.17 \\
\hline
\textit{...a la...} $\leftrightarrow$ \textit{...with ham peppers and chillis...} & 0.17 \\
\hline
\textit{...a la...} $\leftrightarrow$ \textit{...with ham peppers and...} & 0.17 \\
\hline
Total & 1.0 \\
\hline
\textbf{Phrase in v2} & \textbf{Probability}  \\
\hline
\textit{...a\&la...} $\leftrightarrow$ \textit{...with...} & 0.82 \\
\hline
\textit{...a\&la...} $\leftrightarrow$ \textit{...\`a\&la...} & 0.18 \\
\hline
Total & 1.0 \\
\hline
\end{tabular}
\label{tab:newprob}
\end{table}
\normalsize


\begin{table}[htb] \small \renewcommand{\arraystretch}{1.3}
\centering
\caption{Extract from the trained phrase-table, containing standardized translations and probability weights for versions 1, 2 and 3.}
\begin{tabular}{|c|c|c|c|}
\hline
\textbf{Phrase} & \textbf{v1 Prob} & \textbf{v2 Prob} & \textbf{v3 Prob} \\
\hline
\textit{...al...} $\leftrightarrow$ \textit{...with...} & 0.03 & 0.35 & 0.17 \\
\hline
\textit{...al...} $\leftrightarrow$ \textit{...au...} & - & 0.65 & 0.62 \\
\hline
\textit{...en...} $\leftrightarrow$ \textit{...in...} & 0.62 & 0.70 & - \\
\hline
\textit{...en...} $\leftrightarrow$ \textit{...with...} & - & 0.30 & - \\
\hline
\textit{...del...} $\leftrightarrow$ \textit{...of the...} & 0.09 & 1.0 & 1.0 \\
\hline
\textit{...a la...} $\leftrightarrow$ \textit{...with...} & - & 0.82 & 0.77 \\
\hline
\textit{...a la...} $\leftrightarrow$ \textit{...\`a la...} & - & 0.18 & 0.12 \\
\hline
\end{tabular}
\label{tab:defo}
\end{table}
\normalsize

\subsection{Translation accuracy and quality}
\label{ssec:experiments-accuracy}
We tested the accuracy of all versions of our system, as well as that of Google Translate, using our test set.
We chose Google Translate as our benchmark because it is among the most popular popular free machine translation engines, and it is widely used by the general public~\cite{Gaspari_2006, Wan2009, Mathers2010}. Moreover, Google Translate has achieved very good results on several recent machine translation studies~\cite{2009Workshop, Wan2008, Callison-Burch2009}. Note that Google Translate has even been excluded from the last Workshop on Statistical Machine Translation events because of the large margin between Google and many of the other systems. However, it is important to be aware of the limitations of Google Translate. It is not able to select the appropriate context from the input text, and it usually fails to detect idiomatic expressions~\cite{Sheppard2011}. Therefore, it is unlikely to be useful to scientific or medical contexts in the near future~\cite{MOJ2011}.\\
Our test set is a 500 entry list of Spanish restaurant items\footnote{Test data available online: https://redpill.ecn.purdue.edu/$\sim$reddo/resources} containing 2.74 words per entry on average, and with 301 combinations of dishes. All the entries were extracted from different online sources, unrelated to the sources from where we built our corpora.
The accuracy was subjectively determined by the first author of this paper (a native Spanish speaker) (see Table \ref{tab:accuracytest2}).
All incorrect translations in v2 and v3 are due to non-existing words in the phrase-table, or gender and number grammatical errors. Most errors in Google Translate are literally correct but incorrect in a food-related context. For example, \textit{andrajos}, which is a Spanish stew, is translated to \textit{rag}.
The first version (v1) has a slightly better accuracy than Google. 
However, v2 and v3 are both significantly more accurate.
This indicates that translation standardization and n-gram consolidation does indeed increase translation accuracy.


\begin{table}[htb] \small \renewcommand{\arraystretch}{1.3}
\centering
\caption{Translation accuracy, manually evaluated, for versions 1, 2 and 3 of our system, and Google Translate. Top-1 means the first translation is the correct one.}
\begin{tabular}{|c|c|c|c|c|c|}
\hline
\textbf{Engine} & \textbf{Correct} & \textbf{Top-1} & \textbf{Incorrect} & \textbf{Accuracy} & \textbf{Top-1}  \\
\hline
Google & 365 & - & 135 & 73\% & - \\
\hline
v1 & 375 & - & 125 & 75\% & - \\
\hline
v2 & 434 & 416 & 66 & 86.8\% & 83.2\% \\
\hline
v3 & 394 & 375 & 106 & 78.8\% & 75\% \\
\hline
\end{tabular}
\label{tab:accuracytest2}
\end{table}
\normalsize


To fully evaluate the effectiveness and accuracy of our system, one would need to consider the combination of the text translation, the ingredient list, and the picture. Indeed, when seeing the text translation alone, some users may perceive it as incorrect if some words are left unchanged. 
Thus, it is not sufficient to compare merely the text portion of our translation system with a text-based translation tool. 




We conducted a user study of our system to compare our translation and interpretation modules against Google Translate.
Twenty subjects participated in the user study.
Each subject was presented with a mobile device application that showed a list of restaurant menu items in Spanish.
The subject was then shown a form on a desktop computer where they were asked to identify sensitive ingredients for each menu item (alcohol, pork, egg, peanuts, milk, shellfish, wheat).
The answers for each ingredient were either ``Yes'', ``No'' or ``Don't Know''.
Next, the subject was given the English translation (obtained either with Google Translate or with our system) of each of the restaurant menu items on the mobile device, and was given the opportunity to change their answer.
In the case where the translation was obtained with our system, the subject was given images and a list of ingredients to help interpret the text translation (on the mobile device).
After that the subject was again given the opportunity to change their answer.

A total of 33 dishes containing 102 distinct ingredients were used for the experiment.\footnote{Test data available online: https://redpill.ecn.purdue.edu/$\sim$reddo/resources}
Ten of them were chosen at random and used for each subject.
Of this ten we chose a random amount to be translated using Google Translate, and the rest to be translated using our system (first using text only, and subsequently with an image an ingredient list added).

\begin{figure*}[htb]
\begin{center}
\includegraphics[width=\textwidth]{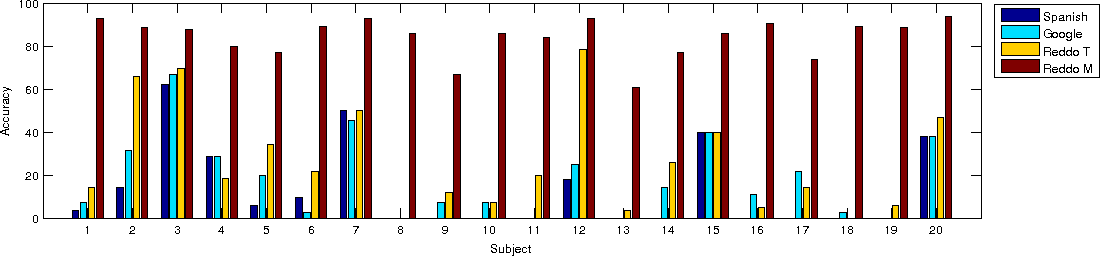}
\caption{Accuracy results of the user study for each subject.}
\label{fig:user-study}
\end{center}
\end{figure*}

Figure \ref{fig:user-study} illustrates the results of the study for each subject, and Table \ref{tab:user-study} shows the average accuracy and standard deviations for each case: Spanish, text translation using Google Translate, text translation using our system (Reddo T), and translation and interpretation (text, lists of ingredients and images) using our system (Reddo M).
While our translation and interpretation system clearly provides more accurate results, it is interesting to note that even the text translation alone is more accurate than Google Translate.

\begin{table}[htb] \small \renewcommand{\arraystretch}{1.3}
\centering
\caption[]{Average accuracies and standard deviations (SD) of the user study for each case.}
\begin{tabular}{|c|c|c|c|c|}
\hline
\textbf{Case} & \textbf{Spanish} & \textbf{Google} & \textbf{Reddo T} & \textbf{Reddo M} \\
\hline
\textbf{Accuracy} & 15.20\% & 18.38\% & 26.59\% & 84.13\% \\
\hline
\textbf{SD} & 26.18\% & 18.58\% & 24.28\% & 9.05\% \\
\hline
\end{tabular}
\label{tab:user-study}
\end{table}

\subsection{Execution time}
\label{ssec:experiments-speed}
We tested the speed of our proposed method (v2) using some entries from our test set. We tested both existing and non-existing items in our phrase-tables. Because the test set entries are relatively short, no significant difference in computation time was observed between entries of differing length. This was also true for n-grams that were not in our phrase-tables.

To put in perspective the execution time of our method (translation standardization and n-gram consolidation), we took the \textit{Spanish-English WMT08 News Commentary} corpora and divided it into subsets of increasing size. Then we used Moses to obtain phrase-tables for each of these subsets. The execution time for each subset was recorded and plotted against the training data size. Table \ref{tab:speedtestB} summarizes the results when using the Spanish input \textit{arroz a la cubana}. Notice that the computation time increases more or less linearly with the corpora size. Notice also that v2 appears to be significantly faster than when training Moses with a corpora of comparable size to ours (1/4\% of WMT08). Thus we believe that our method yields a significant execution time saving with respect to the training set size.

The difference in translation times between v2 and 1/4\% of WMT08 can be partly explained by the difference in n-gram distributions in their respective corpora. Indeed, as can be seen in Table \ref{tab:v2ngramswmt08}, the latter contains significantly more 1-grams, 2-grams and 3-grams. This increases the number of hypothesis that must be considered when translating, thus increasing the execution time. Notice that the automatic consolidation method implemented in v3 is applied once to the entire corpora before the training process, and to each individual user search afterward. The former is the one that affects the translation time, but it takes orders of magnitude less time than the translation itself. Therefore, there is no significant different in translation times between v2 and v3.

\begin{table}[htb] \small \renewcommand{\arraystretch}{1.3}
\centering
\caption{Translation times comparisons, on the desktop, between v2 and the \textit{WMT08} data set, for the Spanish input \textit{arroz a la cubana}.}
\begin{tabular}{|c|c|c|}
\hline
\textbf{Data set} & \textbf{Corpora size} & \textbf{Speed}  \\
\hline
$\frac{1}{4}$ of WMT08 & 100,000 words & 8.5 s \\
\hline
$\frac{1}{8}$ of WMT08 & 50,000 words & 4.5 s \\
\hline
$\frac{1}{16}$ of WMT08 & 25,000 words & 2.5 s \\
\hline
v2 & \textbf{11,067 words} & \textbf{0.5 s} \\
\hline
$\frac{1}{40}$ of WMT08 & 10,000 words & 1.0 s \\
\hline
$\frac{1}{160}$ of WMT08 & 2,500 words & 0.0 s \\
\hline
\end{tabular}
\label{tab:speedtestB}
\end{table}
\normalsize


\begin{table}[htb] \small \renewcommand{\arraystretch}{1.3}
\centering
\caption[]{n-gram counts for version 2 of our system and 1/40\% of the \textit{WMT08 News Commentary} data set.}
\begin{tabular}{c|c|c|c|}
\hline
\multicolumn{1}{|c|}{\textbf{Data set}} & \textbf{1-gram} & \textbf{2-grams} & \textbf{3-grams} \\
\hline
\multicolumn{1}{|c|}{1/40\% WMT08} & 31,241 & 15,578 & 5,650 \\
\hline
\multicolumn{1}{|c|}{v2} & 2,695 & 8,324 & 781 \\
\hline
 & \textbf{-91.37\%} & \textbf{-46.56\%} & \textbf{-86.17\%} \\
\cline{2-4}
\end{tabular}
\label{tab:v2ngramswmt08}
\end{table}

Computation time is one of the most important concerns in mobile devices. Both the loading time and the translation time are relevant. We define the loading time as the time it takes to load the phrase-tables (the trained phrase-table, as well as the six one-to-one phrase-tables), the language model, and the reordering table. This is the time it takes between launching the application and being able to perform the first translation. Table \ref{tab:iphonespeed} shows that the average loading times for the non-memory-mapped (10 seconds) is higher than the memory-mapped (8 seconds) in all five tests in a Second Generation iPod Touch. This was expected, because the memory-mapped case would read from disk on demand, instead of loading the entire phrase-table into memory at the beginning. Note that the numerical results provided by the Moses engine is rounded to the closest integer, so the worst case is always taken (e.g., total time from A to B: 6 seconds, and total time from A to C: 8 seconds $\rightarrow$ total time from B to C: [1.01, 2.99] seconds). Notice that the time period during which the loading times are computed the only difference between v2 and v3 is the memory size of the trained phrase-table. However, this difference is not large enough to be taken into consideration. Moreover, as it can be seen in Table \ref{tab:iphonespeed} the contribution of the trained phrase-table to the total loading time is always minimal.

\begin{table*}[htb] \small \renewcommand{\arraystretch}{1.3}
\centering
\caption[]{Loading times, on the iPod, for the language model (LM), all six one-to-one phrase-tables (PT 1-6), the trained phrase-table (PT) and the reordering table (RT), for the non-memory-mapped and memory-mapped cases, for five tests in the same conditions, in a Second Generation iPod Touch device. All values are the highest possible (worst case).}
\begin{tabular}{c|c|c|c|c|c|c|c|c|c|c|}
\cline{2-11}
 & \multicolumn{5}{c|}{\textbf{Binarized}} & \multicolumn{5}{c|}{\textbf{Non-binarized}} \\
\hline
\multicolumn{1}{|c|}{\textbf{Test}} & \textbf{LM} & \textbf{PT 1-6} & \textbf{PT} & \textbf{RT} & \textbf{TOTAL} & \textbf{LM} & \textbf{PT 1-6} & \textbf{PT} & \textbf{RT} & \textbf{TOTAL} \\  
\hline
\multicolumn{1}{|c|}{1} & 7 s & 1 s & 1 s & 2 s & 8 s & 7 s & 4 s & 1 s & 1 s & 10 s \\
\hline
\multicolumn{1}{|c|}{2} & 6 s & 1 s & 1 s & 3 s & 8 s & 6 s & 5 s & 1 s & 1 s & 10 s \\
\hline
\multicolumn{1}{|c|}{3} & 6 s & 1 s & 1 s & 3 s & 8 s & 6 s & 5 s & 1 s & 1 s & 10 s \\
\hline
\multicolumn{1}{|c|}{4} & 7 s & 1 s & 1 s & 2 s & 8 s & 7 s & 4 s & 1 s & 1 s & 10 s \\
\hline
\multicolumn{1}{|c|}{5} & 7 s & 1 s & 1 s & 2 s & 8 s & 6 s & 5 s & 1 s & 1 s & 10 s \\
\hline
\multicolumn{1}{|c|}{\textbf{AVG}} & \textbf{7 s} & \textbf{1 s} & \textbf{1 s} & \textbf{3 s} & \textbf{8 s} & \textbf{7 s} & \textbf{5 s} & \textbf{1 s} & \textbf{1 s} & \textbf{10 s} \\
\hline
\end{tabular}
\label{tab:iphonespeed}
\end{table*}
\normalsize

We also measured the CPU time needed to translate various menu items after the application being loaded. The translation times for the Spanish food item \textit{arroz a la cubana} in five tests under the same conditions showed almost instantaneous results, with an average of 0.02 seconds. Similar times were obtained for different dishes and word combinations. Table \ref{tab:iphonebintransspeed} shows more results for the memory-mapped case. Note that in the non-memory-mapped case, the phrase-tables would have to be loaded before each translation. In that case, all the translation times in Table \ref{tab:iphonebintransspeed} would be increased by 10 seconds.


\begin{table}[htb] \small \renewcommand{\arraystretch}{1.3}
\centering
\caption{Translation times, on the iPod, for the first (A1) and successive (A1+) translation attempts of the Spanish dish \textit{arroz a la cubana}, using memory-mapped phrase-tables.}
\begin{tabular}{c|c|c|c|c|c|c|}
\cline{2-7}
& \textbf{Test 1} & \textbf{Test 2} & \textbf{Test 3} & \textbf{Test 4} & \textbf{Test 5} & \textbf{Average} \\
\hline
\multicolumn{1}{|c|}{\textbf{A1}} & 2.10 s & 2.06 s & 2.09 s & 2.10 s & 2.12 s & \textbf{2.09 s}\\  
\hline
\multicolumn{1}{|c|}{\textbf{A1+}} & 0.02 s & 0.00 s & 0.02 s & 0.03 s & 0.02 s & \textbf{0.02 s} \\  
\hline
\end{tabular}
\label{tab:iphonebintransspeed}
\end{table}
\normalsize

\subsection{Memory size}
\label{ssec:experiments-memory}
The second version of application, v2, requires 17.52 MB of physical memory on the desktop computer. This includes the main executable, the language model file, the trained phrase-table and the six one-to-one phrase-tables (i.e., without the database of images). On the iPod Touch, the total memory size of the six phrase-tables plus the language model file is 2.6 MB. The size of the entire application, including the executable and the database of images, is 9.56 MB. It is less than in the desktop version, due to the fact that the portable version does not need the SRILM tool that was used by Moses to build the language model; the Moses engine's internal language model is used instead. As previously mentioned, the non-memory-mapped phrase-tables are lighter. However, since the increase on memory size on the memory-mapped case is not a problem, its decrease in translation time is beneficial for the user (see Tables \ref{tab:iphonebintransspeed} and \ref{tab:iphonenonsizes}).


\begin{table}[htb] \small \renewcommand{\arraystretch}{1.3}
\centering
\caption[]{Non-memory-mapped (NMM) and memory-mapped (MM) tables and sizes on the iPod.}
\begin{tabular}{|c|c|c|}
\hline
\textbf{Table} & \textbf{NMM size} & \textbf{MM Size} \\
\hline
phrase-table & 119 KB & 585 KB \\
\hline
reordering-table & 74 KB & 1,576 KB \\
\hline
LM file & 238 KB & 238 KB \\
\hline
\textbf{TOTAL} & \textbf{431 KB} & \textbf{2,399 KB} \\
\hline
\end{tabular}
\label{tab:iphonenonsizes}
\end{table}
\normalsize

The database used for the tests contains 155 images of dishes and ingredients. Its total size is 5.24 MB. Assuming linearity of growth, increasing the database to 1,000 images would increase the memory to 37.82 MB. This is only a 0.46\% of the total 8 GB of internal storage of the lightest model of Second Generation iPod Touch.


\section{Conclusions}
\label{sec:conclusions}

We have proposed a network independent, hand-held system for translation and disambiguation of foreign restaurant menus in real-time.
The system relies on the use of a hand-held portable device, including but not limited to mobile telephones or PDAs.
Our translation module was obtained by modifying Moses, an existing open source statistical machine translation software. In particular, we have proposed a method for modifying a corpora to use in the training process of a Statistical Machine Translation engine. These modifications include translation standardization and n-gram consolidation.
These modifications yield an increase in accuracy and a reduction of computational complexity of the decoding process. They also yield a phrase-table small enough to fit in a hand-held device. \\
We implemented a prototype of our system on an iPod Touch Second Generation for English speakers traveling in Spain. 
Ambiguities and translation errors are reduced through the use of an n-best list along with a browsable database of images of dishes and ingredients combined with ingredient lists and disambiguation dialogs.
The real-time translation is fast (0.09 seconds on average) and the application has a memory size of 9.56 MB, including a small multimedia database of ingredient lists and dish/ingredient images.
In our initial tests, Google Translate yields the correct translation 73\% of the time. In contrast, our system outputs the correct translation in first position 83.2\% of the time. Moreover, the correct translation was within the first three top ranked translations 86.8\% of the time.
By combining our translation system with a database of nutritional and diet specific information, our system can be used as an aid for individuals following special diets. 
More specifically, a profile management system could be used to personalize a diet by flagging ingredients from the database. \\
To test the accuracy of the overall text translation and interpretation system we also conducted a user study focusing on the identification on sensitive ingredients.
Our results show that the translation accuracy of our system outperforms Google Translate.
Indeed the text translation of our system is 8.21\% more accurate than Google Translate on average.
The addition of images and ingredient lists increases the accuracy 65.75\% more accurate than Google Translate, reaching an 84.13\% accuracy.
\\
When appropriate, disambiguation of the translation results would be handled through bilingual dialogs, where information or instructions to be transmitted to the waiter would be suggested by the device.
A current limitation of our system is the need to manually build the database of images and lists of ingredients.
In the future we plan to investigate the use of crowdsourcing systems for this purpose~\cite{Callison2010,Marcus2011}.

\bibliographystyle{IEEEtran}
\bibliography{TD}   

%

\begin{IEEEbiography}
[{\includegraphics[width=1in,height=1.25in,clip,keepaspectratio]{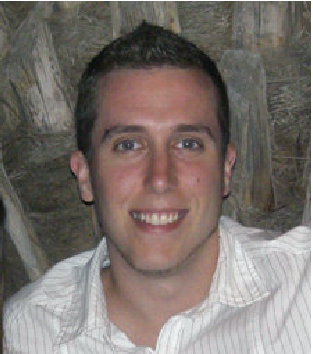}}]
{Albert Parra}
was born in Barcelona, Spain. He received the B.S. degree in Telecommunications Engineering from the Universitat Polit\`ecnica de Catalunya in 2010. He was a visitor scholar and researcher in the Video and Image Processing Laboratory (VIPER) at Purdue University between 2009 and 2010. He received the M.S. degree in Electrical and Computer Engineering from Purdue University in 2011. He is currently pursuing a Ph.D. in Electrical and Computer Engineering at Purdue University under the direction of Prof. Delp and Prof. Boutin.
\end{IEEEbiography}

\begin{IEEEbiography}
[{\includegraphics[width=1in,height=1.25in,clip,keepaspectratio]{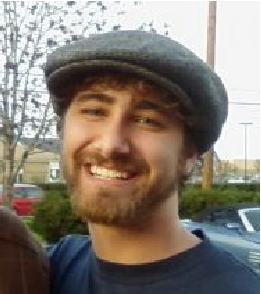}}]
{Andrew W. Haddad}
 received the B.S. in Computer Science from Ball State University, Muncie, IN. He received the M.S. degree in Electrical and Computer Engineering from Purdue University under the direction of Prof. Boutin in the School of Electrical and Computer Engineering.
\end{IEEEbiography}

\begin{IEEEbiography}
[{\includegraphics[width=1in,height=1.25in,clip,keepaspectratio]{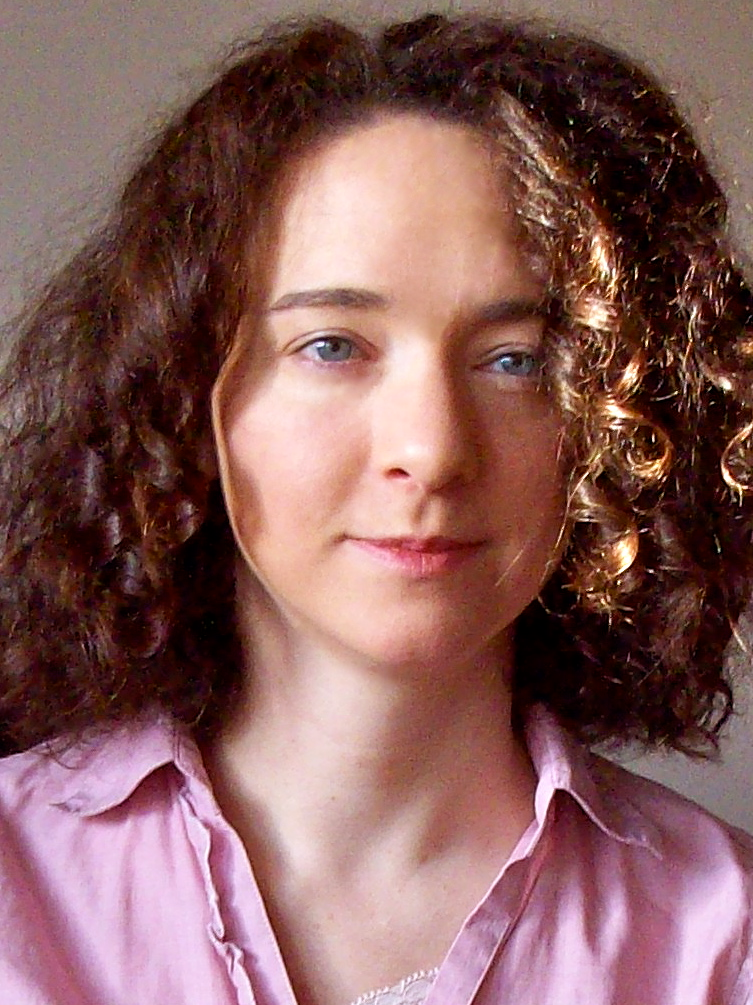}}]
{Mireille Boutin}
 was born in Canada. She received the B.Sc. degree in Physics-Mathematics from the Universit\'e de Montr\'eal, and the Ph.D. degree in Mathematics from the University of Minnesota in Minneapolis. After a post-doctorate at Brown University, followed by a post-doctorate at the Max Planck Institute for Mathematics in the Sciences (Leipzig, Germany), she joined Purdue's School of Electrical and Computer Engineering, where she is now an associate professor.
\end{IEEEbiography}

\begin{IEEEbiography}
[{\includegraphics[width=1in,height=1.25in,clip,keepaspectratio]{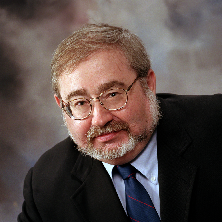}}]
{E.~J.~Delp}
(S'70-M'79-SM'86-F'97) was born in Cincinnati, OH. He received the B.S.E.E. (cum laude) and M.S. degrees from the University of Cincinnati and the Ph.D. degree from Purdue University, West Lafayette, IN. In May 2002, he received an Honorary Doctor of Technology from Tampere University of Technology, Tampere, Finland.
From 1980 to 1984, he was with the Department of Electrical and Computer Engineering, The University of Michigan, Ann Arbor. Since August 1984, he has been with the School of Electrical and Computer Engineering and the School of Biomedical Engineering, Purdue University. He is currently the Charles William Harrison Distinguished Professor of Electrical and Computer Engineering and Professor of Biomedical Engineering. His research interests include image and video compression, multimedia security, medical imaging, multimedia systems, communication, and information theory.
Dr. Delp is a Fellow of the SPIE, a Fellow of the Society for Imaging Science and Technology (IS\&T), and a Fellow of the American Institute of Medical and Biological Engineering. In 2004 he received the Technical Achievement Award from the IEEE Signal Processing Society for his work in image and video compression and multimedia security. In 2008 Dr. Delp received the Society Award from IEEE Signal Processing Society (SPS). This is the highest award given by SPS and it cited his work in multimedia security and image and video compression. In 2009 he received the Purdue College of Engineering Faculty Excellence Award for Research. 
\end{IEEEbiography}

\end{document}